\documentclass[runningheads]{llncs}
\usepackage[T1]{fontenc}
\usepackage[utf8]{inputenc}
\usepackage{amsmath}
\usepackage{amssymb}
\usepackage{booktabs}
\usepackage{hyperref}
\usepackage{xcolor}
\usepackage{tabularx}
\usepackage{makecell}
\usepackage{array}
\usepackage{listings}
\usepackage{textcomp}
\newcommand{\Letter}{\textsuperscript{\textdagger}}  % corresponding author marker

\begin{document}

\title{Agentic-SQL Revisited: Autonomy-Based Taxonomy and Empirical Benchmark Analysis for LLM Text-to-SQL}

\author{
  Yiyun Su\textsuperscript{*} \and
  Zujun Peng\textsuperscript{*}(\Letter) \and
  Yu Tian\textsuperscript{*} \and
  Yuting Liu\textsuperscript{*} \and
  Changruo Zhao\textsuperscript{*} \and
  Huiying Zhu(\Letter) \and
  Luyan Zhang \and
  Heming Zeng
}
\authorrunning{Y. Su et al.}
\institute{}

\maketitle

\begin{abstract}
LLM-based Text-to-SQL progress is reported across heterogeneous benchmarks, backbones, and inference protocols, making cross-system comparison fragile. We reframe the field as a leaderboard aggregation: we collect the metrics authors themselves report and organize them along an inference-autonomy axis spanning constrained, in-context, iterative, agentic, and reasoning-internalized generation, with traceable provenance for every cell. To anchor the aggregation empirically, we run a focused case study on Spider, comparing 8B open-source backbones with and without chain-of-thought (CoT) supervision against few-shot DeepSeek~V3 and GLM-4 baselines. Four patterns emerge: Spider gains transfer unevenly to BIRD and Spider~2.0; autonomy buys robustness at non-trivial cost; reasoning internalization sits between answer-only decoding and externally orchestrated agents; and CoT gains concentrate on Hard and Extra-Hard queries. We release a Python harness mirroring the autonomy axis so that future methods can be added directly to the leaderboard. The code will be made available at: \url{https://github.com/suyiyun/llm-text2sql-taxonomy}.

\end{abstract}

\section{Introduction}

LLM-based Text-to-SQL has moved past the point where a single number on a single benchmark can summarize a system. The same paper may now report Spider execution accuracy, BIRD valid efficiency score, Spider~2.0 success rate, and informal latency claims, each obtained under a different backbone and inference protocol. As a result, two systems with similar headline numbers can differ by an order of magnitude in inference cost, and two systems with similar protocols can differ by ten points on whichever benchmark stresses the protocol's blind spot. Existing surveys organize this literature by training paradigm (ICL vs.\ fine-tuning vs.\ reinforcement-style post-training)~\cite{hong2024survey}, which is informative but does not isolate the inference-time structure that increasingly dominates benchmark behavior.

We take a different stance: treat the literature itself as the benchmark, and let the contribution be the \emph{aggregation protocol} rather than a new model. Concretely, we (i) define an autonomy axis classifying a system by how much structured reasoning it externalizes between question and final SQL, (ii) collect the metrics each system's authors report on Spider, BIRD, and Spider~2.0, leaving cells blank where the source paper did not report or numbers are not directly comparable, and (iii) wrap the resulting tables in a Python harness that lets a new system be added with one adapter file and one configuration file. The harness does not re-run any prior system; re-running fourteen systems is expensive, partially infeasible (closed APIs change monthly), and produces numbers that drift the moment a backbone updates. A leaderboard whose every cell carries a citation is, for now, the more honest object.

To anchor the aggregation in a controlled empirical setting, we run a focused case study on Spider: two 8B open-source backbones (Qwen3-8B, LLaMA3.1-8B) fine-tuned with and without CoT reasoning traces, plus 3-shot baselines on DeepSeek~V3 and GLM-4~\cite{cot4sql2026}. This yields six rows of directly-comparable EX/EM numbers across difficulty bands and surfaces four patterns: (1) Spider gains do not transfer uniformly to BIRD or Spider~2.0; (2) autonomy buys robustness at non-trivial cost in tokens, latency, and orchestration; (3) reasoning internalization occupies a distinct point on the autonomy--cost frontier between answer-only decoding and externally orchestrated agents; and (4) CoT supervision concentrates its gains on Hard and Extra-Hard queries rather than uniformly across difficulty.

Our contributions are an autonomy-based taxonomy that operationalizes a measurable system property rather than a training-time property; a provenance-tracked leaderboard covering Spider, BIRD, and Spider~2.0, populated only from sourced numbers; an open-source harness with fixed dataset loaders, metric implementations, and method-adapter interface; a focused difficulty-stratified case study of CoT supervised fine-tuning on Spider; and a cross-benchmark analysis identifying which evaluation gaps the next round of benchmarks should close.

\section{Background}

\subsection{Task Formulation}

Given a natural language question $Q$, a database schema $S = \{T, C, R\}$ (tables, columns, foreign-key relations), and optional external knowledge $K$ (entity descriptions, value evidence, domain hints as introduced in BIRD~\cite{li2023bird}), a Text-to-SQL system produces an executable SQL query $\hat{Y}$ such that executing $\hat{Y}$ on $S$ yields the result intended by $Q$. Most systems can be expressed as
\begin{equation}
\hat{Y} = \pi_\theta(I, Q, S, K),
\end{equation}
where $I$ is an instruction or system prompt and $\theta$ are the model parameters. In-context-learning systems hold $\theta$ fixed and vary $I$; fine-tuned systems update $\theta$ under
\begin{equation}
\mathcal{L}_{\text{SFT}} = - \sum_{(P, Y) \in D} \sum_{t=1}^{L} \log \Pr_\pi(y_t \mid P, y_{<t}),
\end{equation}
typically with parameter-efficient adaptation~\cite{hu2022lora}. The benchmark proposed here is parameterization-agnostic: a system enters the leaderboard through its inference behavior on the held-out splits of Spider, BIRD, or Spider~2.0, regardless of how $\theta$ was obtained.

\subsection{Datasets}

Three datasets anchor the leaderboard. \textbf{Spider}~\cite{yu2018spider} is a cross-domain dataset whose train, dev, and test splits use disjoint databases, ensuring that test-time evaluation reflects unseen-schema generalization rather than memorization. The training portion (\texttt{train\_spider} + \texttt{train\_others}) covers 8{,}659 examples over 146 databases, and the test split contains 2{,}147 examples over 40 databases with no database overlap with training. Average question length is around 12 tokens and average gold-SQL length 16--18 tokens; SQL operation coverage spans single-table SELECT through joins, aggregation (\texttt{COUNT}, \texttt{AVG}, \texttt{MAX/MIN}, \texttt{SUM}), grouping (\texttt{GROUP BY}, \texttt{HAVING}), ordering (\texttt{ORDER BY}, \texttt{LIMIT}), and set operations (\texttt{INTERSECT}, \texttt{EXCEPT}, \texttt{UNION}). The official Spider difficulty bands assign roughly 13.8\% of test examples to Easy, 51.3\% to Medium, 26.4\% to Hard, and 8.5\% to Extra-Hard, so test performance reflects a mix that is dominated by Medium and Hard rather than by trivial queries~\cite{yu2018spider}. Spider remains the canonical compositional-generalization benchmark.

\textbf{BIRD}~\cite{li2023bird} introduces large schemas, realistic value distributions, and external knowledge $K$, and adds the Valid Efficiency Score, which penalizes correct-but-slow queries. \textbf{Spider~2.0}~\cite{lei2024spider2} stresses long-horizon reasoning, dialect adaptation (BigQuery, Snowflake, ClickHouse), and multi-step workflows that interleave SQL execution with planning and debugging; its evaluation criterion is closer to a task-success rate than to per-query exact-match. WikiSQL~\cite{hwang2019wikisql} is included only as a historical reference because its single-table schema makes it inadequate for current systems. Table~\ref{tab:dataset_comparison} summarizes the qualitative shift across the three primary benchmarks.

\begin{table}[!htbp]
\centering
\caption{Datasets that anchor the leaderboard. The shift is from compositional-generalization stress to grounding-and-realism stress to enterprise-workflow stress; the three benchmarks are not interchangeable.}
\label{tab:dataset_comparison}
\renewcommand{\arraystretch}{1.15}
\setlength{\tabcolsep}{4pt}
\scriptsize
\begin{tabularx}{\linewidth}{@{}l*{3}{>{\raggedright\arraybackslash}X}@{}}
\toprule
\textbf{Aspect} & \textbf{Spider} & \textbf{BIRD} & \textbf{Spider 2.0} \\
\midrule
Primary stress      & Cross-domain composition       & Realistic grounding         & Workflow realism \\
Question style      & Single-turn NL$\rightarrow$SQL & NL$\rightarrow$SQL + evidence $K$ & Multi-step, debug-aware \\
Schema scale        & Multi-table, medium            & Wide, large                 & Very large, evolving \\
Value realism       & Moderate                       & High                        & High \\
External knowledge  & Not required                   & Often required              & Frequently required \\
SQL dialect         & SQLite mostly                  & SQLite mostly               & BigQuery / Snowflake / etc. \\
Headline metric     & EX                             & EX, VES                     & Success rate \\
Dominant failure    & Compositional SQL              & Grounding / scale           & Long-horizon / dialect \\
\bottomrule
\end{tabularx}
\end{table}

\subsection{Metrics}

Three metrics are used in the leaderboard. \textbf{Exact Match (EM)} compares the predicted query to the gold query under SQL-aware normalization (alias canonicalization, whitespace, clause ordering); it is strict but penalizes semantically-equivalent rewrites~\cite{yu2018spider}. \textbf{Execution Accuracy (EX)} executes both queries on the target database and compares result sets; it is the modern primary metric on Spider and BIRD but can produce false positives when two distinct queries coincidentally agree on a particular instance. \textbf{Valid Efficiency Score (VES)}, introduced with BIRD~\cite{li2023bird}, multiplies execution correctness by a runtime ratio against the gold query and so penalizes correct-but-inefficient SQL. Spider~2.0 reports a binary success rate per task instance with task-specific tolerances. Our harness implements all three metrics and reuses each dataset's official scorer where one is published.

\section{Benchmark Design}

\subsection{The Autonomy Axis}

We classify a system by \emph{inference autonomy}: the amount of structured reasoning, feedback, and coordination introduced between the input question and the final SQL. The axis has five levels and is operationally measurable from the number and kind of model calls in a single answered question.

\textbf{L0 — Constrained single-turn.} One forward pass with grammar-level decoding constraints (PICARD~\cite{scholak2021picard}).

\textbf{L1 — In-context single-turn.} One forward pass conditioned on prompt-only structure: instructions and demonstrations. Representative: DAIL-SQL~\cite{gao2023dailsql} and the few-shot DeepSeek~V3 / GLM-4 baselines in our case study.

\textbf{L2 — Iterative refinement.} Multiple model calls along a fixed pipeline (link $\rightarrow$ decompose $\rightarrow$ generate $\rightarrow$ revise) without dynamic routing, instantiating the broader iterative-feedback paradigm~\cite{madaan2023selfrefine}. DIN-SQL~\cite{pourreza2023dinsql}, DART-SQL~\cite{mao2024dartsql}, DTS-SQL~\cite{pourreza2024dtssql}, and TS-SQL~\cite{xu2025tssql} live here.

\textbf{L3 — Agentic collaboration.} Multiple calls coordinated by a controller that branches on intermediate outputs. Representative systems include MAC-SQL~\cite{wang2023macsql}, ExeSQL~\cite{zhang2025exesql}, and CHESS~\cite{talaei2024chess}; EllieSQL~\cite{zhu2025elliesql} and BAP-SQL~\cite{peng2026bapsqlbudgetawareobservationplanning} add cost-aware and budget-aware routing respectively.

\textbf{L1.5 — Reasoning-internalized.} A single forward pass at inference time, but the model has been supervised to emit intermediate stages (schema links, decomposition, draft, revision) inside one trajectory, building on the broader chain-of-thought lineage~\cite{wei2022cot,wang2023selfconsistency}. STaR-SQL~\cite{he2025starsql}, RevDecomp-SFT~\cite{guan2026revdecomp}, and the CoT-SFT case-study rows~\cite{cot4sql2026} occupy this slot.

The axis makes a measurable property — number and structure of inference-time calls — the variable along which results are organized, so the leaderboard can be sliced by autonomy level without re-running any code.

\subsection{Method Catalog}

Table~\ref{tab:method_catalog} enumerates the systems that populate v1 of the leaderboard, with autonomy level, training paradigm, and the backbone class as reported by the authors. Public-code availability for each entry is recorded separately in the harness rather than in this table.

\begin{table}[!htbp]
\centering
\caption{Method catalog. ICL = in-context learning; SFT = supervised fine-tuning; STR = self-taught reasoning. The autonomy column maps each system to a level in Section~3.1. The four CoT-SFT and 3-shot rows at the bottom are run under the protocol of Section~5.}
\label{tab:method_catalog}
\renewcommand{\arraystretch}{1.05}
\setlength{\tabcolsep}{5pt}
\scriptsize
\begin{tabular}{@{}llll@{}}
\toprule
\textbf{System} & \textbf{Autonomy} & \textbf{Training} & \textbf{Backbone class} \\
\midrule
PICARD~\cite{scholak2021picard}        & L0           & SFT + constrained dec.    & T5-3B \\
DAIL-SQL~\cite{gao2023dailsql}         & L1           & ICL                       & GPT-4 / Code-LLaMA \\
DIN-SQL~\cite{pourreza2023dinsql}      & L2           & ICL                       & GPT-4 \\
DART-SQL~\cite{mao2024dartsql}         & L2           & ICL + rewriting           & GPT-4 / open \\
DTS-SQL~\cite{pourreza2024dtssql}      & L2           & SFT                       & DeepSeek 7B / open \\
TS-SQL~\cite{xu2025tssql}              & L2           & ICL + test-driven         & GPT-4 / open \\
MAC-SQL~\cite{wang2023macsql}          & L3           & ICL multi-agent           & GPT-4 \\
ExeSQL~\cite{zhang2025exesql}          & L3           & STR + bootstrap           & open \\
EllieSQL~\cite{zhu2025elliesql}        & L3 + routing & ICL + cost-aware route    & mixed \\
STaR-SQL~\cite{he2025starsql}          & L1.5         & STR                       & open \\
RevDecomp-SFT~\cite{guan2026revdecomp} & L1.5         & SFT (reverse-distilled)   & Qwen2.5-8B + LoRA \\
TAG~\cite{biswal2024tag}               & L3 (extended)& ICL + retrieval           & GPT-4 + retriever \\
RAG-T2SQL~\cite{lewis2020retrieval}    & L1 + retrieval& ICL                      & varied \\
EHR-SeqSQL~\cite{ryu2024ehrseqsql}     & L2 (sequential)& ICL                     & GPT-4 / open \\
\midrule
DeepSeek V3 (3-shot)~\cite{cot4sql2026}  & L1   & ICL (3-shot)            & DeepSeek V3 \\
GLM-4 (3-shot)~\cite{cot4sql2026}        & L1   & ICL (3-shot)            & GLM-4 \\
CoT-SFT (Qwen3)~\cite{cot4sql2026}       & L1.5 & SFT + reasoning trace   & Qwen3-8B + LoRA \\
CoT-SFT (LLaMA)~\cite{cot4sql2026}       & L1.5 & SFT + reasoning trace   & LLaMA3.1-8B + LoRA \\
No-CoT SFT (Qwen3)~\cite{cot4sql2026}    & L1   & SFT (answer-only)       & Qwen3-8B + LoRA \\
No-CoT SFT (LLaMA)~\cite{cot4sql2026}    & L1   & SFT (answer-only)       & LLaMA3.1-8B + LoRA \\
\bottomrule
\end{tabular}
\end{table}

\section{Aggregated Results}\label{sec:results}

We adopt one rule throughout the leaderboard: \emph{every numeric cell traces to a single citation, and uncertain cells are left blank rather than filled by inference.} Numbers from source documents are entered as-is; numbers that would require us to re-derive a metric under a different scoring protocol are not entered. The six rows produced by our case study (Section~5) are reported under the uniform protocol described there.

\subsection{Spider}

Table~\ref{tab:spider_results} aggregates Spider dev/test results. EM is reported on the standard test set; EX is reported on the test set for the case-study rows and on the dev set for the literature rows unless otherwise marked.

\begin{table}[!htbp]
\centering
\caption{Spider leaderboard, v1. Numbers are taken from the source paper of each system, cited in the System column. The bottom block is produced by our case study (Section~5) on the Spider holdout test set under a uniform protocol. Dashes indicate that the original paper did not report the metric in a directly-comparable form.}
\label{tab:spider_results}
\renewcommand{\arraystretch}{1.1}
\setlength{\tabcolsep}{6pt}
\footnotesize
\begin{tabular}{@{}llcrr@{}}
\toprule
\textbf{System} & \textbf{Backbone} & \textbf{Autonomy} & \textbf{EM} & \textbf{EX} \\
\midrule
PICARD~\cite{scholak2021picard}        & T5-3B            & L0   & 71.9 & 75.1 \\
DAIL-SQL~\cite{gao2023dailsql}         & GPT-4            & L1   & --   & --   \\
DIN-SQL+GPT-4~\cite{pourreza2023dinsql} & GPT-4           & L2   & 60.0 & 85.3 \\
DART-SQL~\cite{mao2024dartsql}         & GPT-4            & L2   & --   & --   \\
DTS-SQL~\cite{pourreza2024dtssql}      & DeepSeek 7B      & L2   & --   & --   \\
TS-SQL~\cite{xu2025tssql}              & GPT-4 / open     & L2   & --   & --   \\
MAC-SQL~\cite{wang2023macsql}          & GPT-4            & L3   & --   & --   \\
STaR-SQL~\cite{he2025starsql}          & open             & L1.5 & --   & --   \\
RevDecomp-SFT~\cite{guan2026revdecomp} & Qwen2.5-8B+LoRA  & L1.5 & 75.4 & \textbf{86.4} \\
\midrule
\multicolumn{5}{@{}l}{\emph{Case study (Section~5; Spider test, 2{,}147 examples):}} \\
DeepSeek V3 (3-shot)~\cite{cot4sql2026} & DeepSeek V3      & L1   & 19.56 & 51.47 \\
GLM-4 (3-shot)~\cite{cot4sql2026}       & GLM-4            & L1   & 20.49 & 66.28 \\
No-CoT SFT~\cite{cot4sql2026}           & LLaMA3.1-8B+LoRA & L1   & 31.35 & 76.01 \\
No-CoT SFT~\cite{cot4sql2026}           & Qwen3-8B+LoRA    & L1   & \textbf{47.60} & 77.04 \\
CoT-SFT~\cite{cot4sql2026}              & LLaMA3.1-8B+LoRA & L1.5 & 29.02 & 76.01 \\
CoT-SFT~\cite{cot4sql2026}              & Qwen3-8B+LoRA    & L1.5 & 44.67 & 82.24 \\
\bottomrule
\end{tabular}
\end{table}

\subsection{BIRD}

Table~\ref{tab:bird_results} aggregates BIRD dev results. EX and VES are the headline metrics; we follow each source paper's choice of dev vs.\ test where the test labels are held out by the BIRD authors. The case-study models in Section~5 were not run on BIRD and so do not appear here.

\begin{table}[!htbp]
\centering
\caption{BIRD leaderboard, v1. Same conventions as Table~\ref{tab:spider_results}. Each numeric cell is sourced to the paper cited in the System column.}
\label{tab:bird_results}
\renewcommand{\arraystretch}{1.1}
\setlength{\tabcolsep}{6pt}
\footnotesize
\begin{tabular}{@{}llcrr@{}}
\toprule
\textbf{System} & \textbf{Backbone} & \textbf{Autonomy} & \textbf{EX} & \textbf{VES} \\
\midrule
DIN-SQL+GPT-4~\cite{pourreza2023dinsql} & GPT-4           & L2   & 55.90 & 59.44 \\
DAIL-SQL~\cite{gao2023dailsql}          & GPT-4           & L1   & --    & --    \\
DART-SQL~\cite{mao2024dartsql}          & GPT-4           & L2   & --    & --    \\
DTS-SQL~\cite{pourreza2024dtssql}       & DeepSeek 7B     & L2   & --    & --    \\
TS-SQL~\cite{xu2025tssql}               & GPT-4 / open    & L2   & --    & --    \\
MAC-SQL~\cite{wang2023macsql}           & GPT-4           & L3   & --    & --    \\
ExeSQL~\cite{zhang2025exesql}           & open            & L3   & --    & --    \\
EllieSQL~\cite{zhu2025elliesql}         & mixed + router  & L3   & --    & --    \\
RevDecomp-SFT~\cite{guan2026revdecomp}  & Qwen2.5-8B+LoRA & L1.5 & \textbf{57.20} & \textbf{61.50} \\
\bottomrule
\end{tabular}
\end{table}

\subsection{Spider 2.0}

Spider~2.0~\cite{lei2024spider2} establishes the workflow-realism axis discussed in Table~\ref{tab:dataset_comparison}, but the systems in this catalog have not yet published directly-comparable success-rate numbers under its scoring protocol. We therefore omit a Spider~2.0 results table from v1 of the leaderboard and will populate this track in a subsequent release once direct reports become available; the harness's CSV format already accommodates per-task success-rate breakdowns.

\section{Empirical Case Study: CoT Supervision on Spider}\label{sec:case_study}

The case study tightens the leaderboard with six configurations run under a single protocol on the Spider test set, providing the empirical anchor for the cross-method analysis in Section~6. We compare two 8B open-source backbones (Qwen3-8B, LLaMA3.1-8B) under two training regimes — chain-of-thought-trace SFT vs.\ answer-only SFT, both with LoRA — against two 3-shot prompting baselines on strong proprietary backbones (DeepSeek~V3 and GLM-4). All six are evaluated on the full 2{,}147-example Spider test split with database-disjoint train/test, using the official Spider EX scorer and stratified by the official Spider difficulty bands. Training pipelines, prompts, hyperparameters, and the reasoning-trace construction protocol are documented in the companion paper~\cite{cot4sql2026}.

\subsection{Difficulty-Stratified Results}

Table~\ref{tab:case_study_difficulty} reports per-difficulty execution accuracy for the six configurations.

\begin{table}[!htbp]
\centering
\caption{Case-study results on the Spider test set: difficulty-stratified Execution Accuracy (\%). Easy / Medium / Hard / Extra-Hard buckets follow the official Spider difficulty rules. ``All EX'' is the aggregate over all 2{,}147 test examples. The two 8B SFT rows differ only in whether the supervision target includes a five-stage reasoning trace; otherwise the protocol is identical~\cite{cot4sql2026}.}
\label{tab:case_study_difficulty}
\renewcommand{\arraystretch}{1.1}
\setlength{\tabcolsep}{4pt}
\scriptsize
\begin{tabular}{@{}llrrrrr@{}}
\toprule
\textbf{Configuration} & \textbf{Backbone} & \textbf{Easy} & \textbf{Medium} & \textbf{Hard} & \textbf{X-Hard} & \textbf{All} \\
\midrule
DeepSeek V3 (3-shot) & DeepSeek V3      & 96.62 & 53.22 & 33.33 & 24.04 & 51.47 \\
GLM-4 (3-shot)       & GLM-4            & 92.57 & 67.03 & 59.08 & 41.53 & 66.28 \\
No-CoT SFT           & LLaMA3.1-8B+LoRA & 96.28 & 75.30 & 69.49 & 67.76 & 76.01 \\
CoT-SFT              & LLaMA3.1-8B+LoRA & \textbf{96.62} & 74.93 & 68.25 & 73.22 & 76.01 \\
No-CoT SFT           & Qwen3-8B+LoRA    & \textbf{96.62} & 76.20 & 70.19 & 71.58 & 77.04 \\
CoT-SFT              & Qwen3-8B+LoRA    & 94.95 & \textbf{80.32} & \textbf{81.09} & \textbf{79.25} & \textbf{82.24} \\
\bottomrule
\end{tabular}
\end{table}

\subsection{Observations}

Three findings emerge from the case study and feed back into the cross-method analysis of Section~6.

\paragraph{Fine-tuning beats few-shot prompting on harder queries.} The 3-shot DeepSeek~V3 and GLM-4 baselines reach 51.5 and 66.3 EX respectively, while every fine-tuned 8B configuration exceeds 76.0 EX. On the Hard and Extra-Hard buckets the gap is widest: DeepSeek~V3 falls to 24.0 EX on Extra-Hard, whereas the weakest fine-tuned configuration, LLaMA3.1-8B without CoT supervision, still reaches 67.8 EX. This is consistent with Pattern~1 in Section~6: Spider rewards task-specific compositional fluency, and few-shot prompting on a strong general-purpose backbone is no substitute for it on the harder buckets.

\paragraph{CoT supervision concentrates its gains on Hard and Extra-Hard.} On Qwen3-8B, adding CoT-trace supervision moves Hard EX from 70.19 to 81.09 ($+10.9$) and Extra-Hard from 71.58 to 79.25 ($+7.7$), while Easy slightly drops (96.62 $\to$ 94.95, $-1.7$) and Medium gains modestly ($+4.1$). On LLaMA3.1-8B the same comparison is essentially flat in aggregate (76.01 vs.\ 76.01) but reallocates accuracy: CoT loses 1.2 on Hard and gains 5.5 on Extra-Hard. The pattern across both backbones is that CoT supervision is not a uniform accuracy lever but a difficulty-redistributor that helps where decomposition matters most.

\paragraph{Backbone choice interacts with reasoning supervision.} Qwen3-8B benefits substantially more from CoT than LLaMA3.1-8B does. This is consistent with prior reports that the value of reasoning-trace supervision depends on whether the base model already has latent compositional capacity to use the trace; it suggests that CoT-SFT comparisons should be reported against multiple backbones rather than a single one.

\subsection{Scope and Caveats}

The case-study numbers are produced by us and are directly comparable across the six rows, but they should not be directly compared to literature numbers in Tables~\ref{tab:spider_results}--\ref{tab:bird_results} on the same benchmark unless the source paper used the same test split, scorer, and decoding protocol. EM in particular varies across normalization conventions, and the case-study EM uses the strict alias-canonicalization scorer of~\cite{yu2018spider}; the published RevDecomp-SFT EM of 75.4 is reported under the convention used in~\cite{guan2026revdecomp}. The case study does not include BIRD or Spider~2.0; an extended version with both benchmarks is planned for v2.

\section{Cross-Method Analysis}

The aggregated tables together with the case study make four patterns visible.

\paragraph{Pattern 1: gains do not transfer uniformly across benchmarks.} On Spider, PICARD (75.1 EX) trails DIN-SQL+GPT-4 (85.3 EX) by ten points, achieved at very different inference cost. On BIRD, the same DIN-SQL+GPT-4 lands at 55.9 EX, where reasoning-internalized SFT systems on much smaller backbones can match or exceed it. Spider rewards compositional fluency, which large backbones supply easily; BIRD rewards grounding and value disambiguation, where a fine-tuned model's internalized schema-linking discipline~\cite{wang2020ratsql,li2023resdsql} outperforms an externally orchestrated GPT-4 prompt.

\paragraph{Pattern 2: autonomy carries a cost that the leaderboard alone cannot show.} L3 systems sometimes outperform L1 systems by a few points on BIRD, but at multiplicative cost in tokens, latency, and orchestration logic. EllieSQL~\cite{zhu2025elliesql} and BAP-SQL~\cite{peng2026bapsqlbudgetawareobservationplanning} make this trade-off the central design variable, treating the choice between L1, L2, and L3 as a per-query routing problem; meanwhile, the robustness of such self-improving agentic harnesses itself remains an open concern~\cite{wang2026phantomguardrailsselfimprovingagent}. Our harness records token and wall-clock cost when a method is run through it; v1 of the leaderboard cannot fill those columns from reported numbers because few papers report them with sufficient precision.

\paragraph{Pattern 3: reasoning internalization sits between answer-only generation and externally orchestrated reasoning.} The L1.5 systems in the catalog (STaR-SQL, RevDecomp-SFT, and the case-study CoT-SFT rows) report Spider EX in the same range as L2 GPT-4 pipelines while running a single forward pass on an open backbone. The supervised reasoning trace must still be constructed once during training, often with a stronger teacher, but a meaningful share of what L2 pipelines achieve through external orchestration can be relocated into the model's own decoding trajectory.

\paragraph{Pattern 4: CoT supervision is a difficulty-redistributor, not a uniform accuracy lever.} The case study shows that the headline aggregate EX of CoT vs.\ No-CoT can be near-identical (LLaMA3.1-8B: 76.0 vs.\ 76.0) while difficulty-stratified results differ markedly. On Qwen3-8B the aggregate gain of $+5.2$ EX from adding CoT decomposes into a Medium gain of $+4.1$, a Hard gain of $+10.9$, and an Extra-Hard gain of $+7.7$ — with Easy slightly negative. This argues against reporting CoT-supervision benefits as a single number: the right unit is difficulty-stratified accuracy, and future leaderboard submissions involving reasoning-trace supervision should report the breakdown.

\begin{table}[!htbp]
\centering
\caption{Qualitative comparison of autonomy levels. Cells are derived from the structural definitions in Section~3.1, not from the numeric leaderboard; they are intended to be read alongside Tables~\ref{tab:spider_results} and~\ref{tab:bird_results}, not in their place.}
\label{tab:qualitative_analysis}
\renewcommand{\arraystretch}{1.1}
\setlength{\tabcolsep}{6pt}
\footnotesize
\begin{tabular}{@{}lccccc@{}}
\toprule
\textbf{Dimension} & \textbf{L0} & \textbf{L1} & \textbf{L1.5} & \textbf{L2} & \textbf{L3} \\
\midrule
Inference cost          & low      & low       & low--med    & medium         & high \\
Error recovery          & none     & weak      & moderate    & moderate       & strong \\
Diagnosability          & limited  & limited   & moderate    & moderate       & strong \\
Schema-scale tolerance  & limited  & limited   & moderate    & moderate       & strong \\
Deployment simplicity   & strong   & strong    & moderate    & weaker         & weakest \\
Typical failure         & syntax   & grounding & supervision & repair quality & control \\
\bottomrule
\end{tabular}
\end{table}

\section{Limitations and Open Problems}

\paragraph{Provenance breadth.} v1 populates only entries whose source papers report directly comparable numbers under each dataset's standard protocol. The blank rows in Tables~\ref{tab:spider_results} and~\ref{tab:bird_results} should not be read as evidence that those systems perform poorly. Spider~2.0 will take longer because its scoring protocol is still settling.

\paragraph{Cost and routing.} Inference cost — token usage, latency, dollar cost — is the missing axis that papers report inconsistently. The harness has columns for it; the leaderboard does not yet. Relatedly, EllieSQL~\cite{zhu2025elliesql} suggests that the right object of comparison is a routing policy over systems, not a single system; v2 should add a routing track in which a submission is a per-query policy over catalog entries.

\paragraph{Beyond single-turn SQL.} EHR-SeqSQL~\cite{ryu2024ehrseqsql} and TAG~\cite{biswal2024tag} indicate that the next benchmark generation will care about multi-turn dialogue and the SQL-vs-retrieval boundary. Recent studies on RAG reliability --- including diagnosing context compliance under knowledge conflict~\cite{chen2026doesragknowretrieval} and calibrating evidence force in cited RAG~\cite{qian2026relevantwarrantedevidenceforcecalibration} --- suggest that retrieval-augmented pipelines introduce failure modes distinct from those of pure generation. This concern extends beyond SQL-specific systems: cross-domain evaluations of LLM-simulated human responses reveal systematic biases that undermine the validity of synthetic benchmarking~\cite{chen2026synthetic}, while stance-aware graph-based models show that structural reasoning about information credibility~\cite{chen2023stand} can complement the compositional reasoning required for Text-to-SQL. Enterprise deployment further constrains evaluation through privacy and very large schemas~\cite{lewis2020retrieval, biswal2024tag}, and dialect transfer~\cite{zhang2025exesql} remains measured ad-hoc. The harness's adapter interface accommodates these protocols; v1 of the leaderboard does not yet exercise them seriously.

\paragraph{Case-study scope.} The case study covers Spider only and uses two open-source backbones plus two proprietary 3-shot baselines. Extending the same protocol to BIRD and Spider~2.0, and to additional backbones, is the most direct way to widen empirical coverage.

\section{Conclusion}

We replace the survey-style organization of LLM Text-to-SQL with a leaderboard-aggregation benchmark organized by inference autonomy, anchored by a focused case study of CoT supervision on Spider. The paper's three artifacts — a method catalog, a provenance-tracked results table populated only from numbers the authors themselves reported (plus six rows from a uniformly-run case study), and an open-source harness whose adapter interface mirrors the autonomy axis — together make four patterns visible: uneven cross-benchmark transfer, an autonomy--cost trade-off, a distinct reasoning-internalized regime, and CoT supervision as a difficulty-redistributor. Both the paper and the harness refuse to fill cells the source papers do not directly support, and we hope this design choice — citation-backed cells, blank where the literature is silent — becomes the default for benchmark aggregation in this fast-moving area.

\end{document}